\documentclass[11pt,a4paper]{article}
\usepackage{times,latexsym}
\usepackage{url}
\usepackage[T1]{fontenc}

\usepackage[acceptedWithA]{tacl2021v1}

\usepackage{xspace,mfirstuc,tabulary}

\newif\iftaclinstructions
\taclinstructionsfalse 
\iftaclinstructions
\renewcommand{\confidential}{}
\renewcommand{\anonsubtext}{(No author info supplied here, for consistency with
TACL-submission anonymization requirements)}
\newcommand{\instr}
\fi

\iftaclpubformat 

\else

\fi

\usepackage{times}
\usepackage{latexsym}
\usepackage{booktabs}

\usepackage[T1]{fontenc}

\usepackage[utf8]{inputenc}

\usepackage{microtype}

\usepackage{enumitem}

\usepackage{inconsolata}

\usepackage{graphicx}
\usepackage{soul}
\usepackage{subcaption}

\usepackage{graphicx}
\usepackage{caption}
\usepackage{subcaption}
\usepackage{algorithm}
\usepackage{algorithmic}
\usepackage{booktabs}
\usepackage{dsfont}
\usepackage{multirow}
\usepackage{amsmath}
\usepackage{makecell}

\title{\textit{Self-Consistency Falls Short!} \\ The Adverse Effects of Positional Bias on Long-Context Problems}

\author{
  Adam Byerly
  \and
  Daniel Khashabi
  \\ \ \\
  Johns Hopkins University, USA
  \\
  \texttt{abyerly2@jhu.edu, danielk@cs.jhu.edu}
}

\date{}

\usepackage[most]{tcolorbox}
\usepackage{xparse}
\AtBeginEnvironment{tcolorbox}{\small}
\definecolor{purple}{rgb}{0.5,0,1}
\definecolor{dcyan}{rgb}{0.2,0.6,0.5}
\definecolor{gold}{rgb}{1.0, 0.84, 0.0}

\usepackage[dvipsnames]{xcolor}
\usepackage{placeins}

\begin{document}

\maketitle

\begin{abstract}
Self-consistency (SC) improves the performance of large language models (LLMs) across various tasks and domains that involve short content. However, does this support its effectiveness for long-context problems?

We challenge the assumption that SC's benefits generalize to long-context settings, where LLMs often struggle with position bias---the systematic over-reliance on specific context regions---which hinders their ability to utilize information effectively from all parts of their context. Through comprehensive experimentation with varying state-of-the-art models, tasks, and SC formulations, we find that SC not only fails to improve but actively degrades performance on long-context tasks. This degradation is driven by persistent position bias, which worsens with longer context lengths and smaller model sizes but remains invariant to prompt format or task type. Unlike short-context tasks, where SC diversifies reasoning paths, long-context SC amplifies positional errors. These comprehensive results provide valuable insight into the limitations of current LLMs in long-context understanding and highlight the need for more sophisticated approaches.
\end{abstract}

\section{Introduction}
\label{sec:Introduction}
Large Language Models (LLMs) have shown remarkable versatility in performing various tasks through prompting \citep[inter alia]{brown2020language}. However, these models exhibit various forms of brittleness across tasks~\citep{mishra2022reframing, frieder2024mathematical, mirzadeh2025gsmsymbolic, wu2024reasoning}, including catastrophic failures on simple problems easily solvable by humans \citep{nezhurina2024alice}. To improve reliability, \textit{self-consistency} (SC)~\citep{wang2023selfconsistency} has emerged as a powerful strategy that aggregates multiple sampled responses to mitigate failures. SC has been highly effective in \textit{short} tasks (e.g., $\le100$ tokens), but its ability to handle \textit{long} contexts (e.g., $\ge10K$ tokens) remains underexplored. As real-world applications increasingly demand processing lengthy inputs---such as legal analysis, medical diagnostics, or scientific literature review---understanding SC's scalability is crucial for developing robust solutions. This raises a critical question: Does SC's benefit scale with context length, or do longer contexts introduce challenges that inherently alter its effectiveness?

\begin{figure*}[t!]
    \centering
    \includegraphics[trim={0.3cm 0.5cm 0.3cm 0.5cm},clip,width=0.98\linewidth]{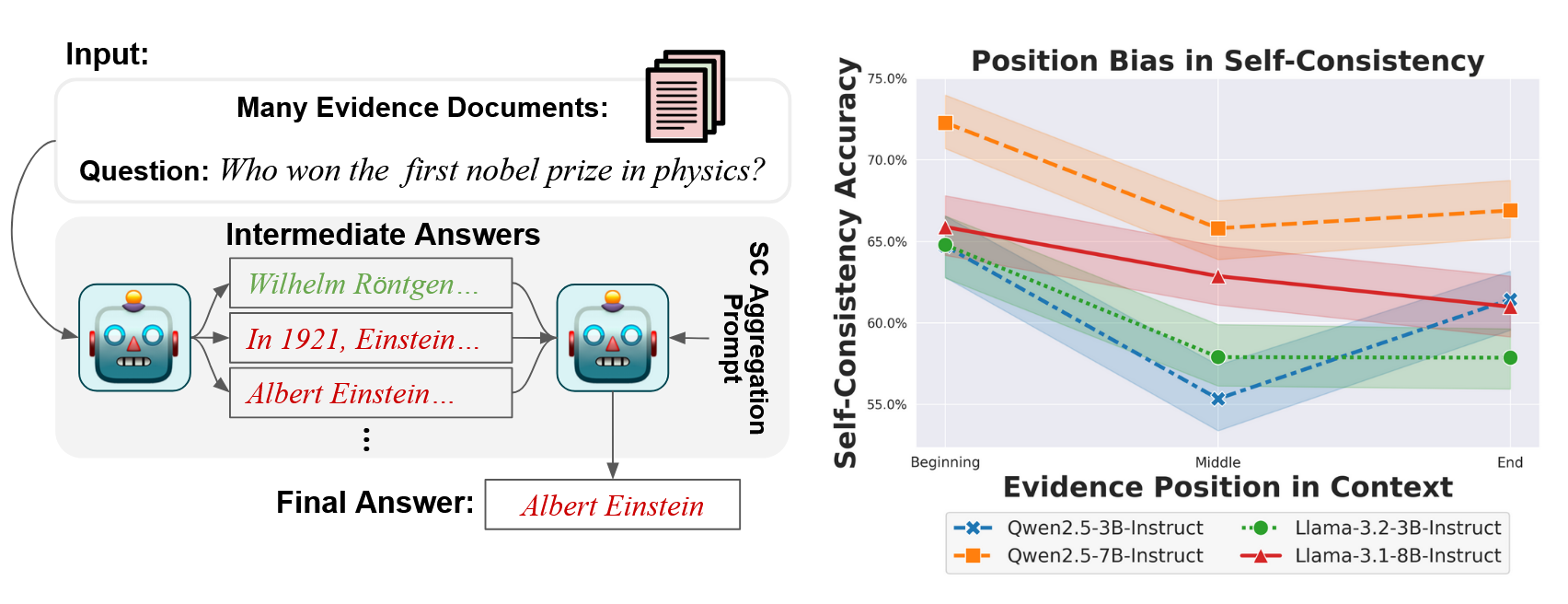}
    \caption{Schematic of self-consistency in long-context, ``needle-in-a-haystack'' scenarios.
    Input consists of a query and multiple evidence documents, one of which contains the correct answer, with a model generating diverse intermediate answers via stochastic sampling (non-zero temperature).
    However, aggregation yields incorrect answers due to position bias, highlighting a key challenge in long-context reasoning.
    Sampling from a model with inherent position bias \textbf{amplifies rather than mitigates errors}, as all samples inherit the same structural biases, violating SC's core assumption of error independence.}
    \label{fig:teaser}
\end{figure*}

Fundamentally, SC assumes errors in individual samples are independent. However, in long-context tasks, systematic position biases induce \textit{correlated errors}, violating SC's core assumption. While prior work \citep{wang2023primacy, zheng2023large, liu2024lost} has documented position bias as a standalone challenge in long-context reasoning, our study provides a diagnostic analysis of how these biases interact with self-consistency, compounding errors through correlated sampling rather than resolving them. We hypothesize that because SC aggregates multiple responses from the same biased model, it reinforces these correlated errors, ultimately amplifying position bias (Figure~\ref{fig:teaser}).

To test this hypothesis and explore the broader implications of SC in long-context scenarios, we conduct extensive experiments across state-of-the-art models (GPT-4o \citep{openai2024gpt}, LLaMA-3.3-70B \citep{grattafiori2024llama}, and Qwen-2.5-72B \citep{yang2024qwen2}) and their smaller variants, evaluating nine diverse long-context tasks spanning summarization, question answering, multi-hop reasoning, and controlled position experiments. Our comprehensive evaluation combines traditional metrics with LLM-based assessment and explores multiple SC implementations, including: majority voting \citep{wang2023selfconsistency} for tasks with discrete answer choices, universal SC \citep{chen2024universal} for open-ended generation tasks, and soft SC \citep{wang2024soft} as an alternative open-ended generation aggregation strategy.

\paragraph{Contributions.}
We provide the first systematic study of self-consistency in long-context settings, revealing three key findings from across 651 carefully designed experiments:
\begin{enumerate}[label=(\roman*)]
    \item \textbf{Failure of SC in long-contexts:} SC consistently degrades performance across models and tasks. This degradation persists across synthetic and real-world tasks, challenging the assumption that SC can universally improve LLM reliability.
    \item \textbf{Robustness to metric and implementation:} 
    SC's long-context limits are confirmed across multiple metrics, aggregation methods, and SC parameterizations.
    \item \textbf{Mechanistic role of position bias:} Through controlled experiments, we show that SC amplifies position bias in both retrieval and reasoning steps. Varying prompt structure and SC configuration reveals that SC's failures stem from \textit{correlated positional errors}.
\end{enumerate}
These insights challenge the assumption that SC universally improves performance and provides guidance for developing more robust approaches to long-context tasks. Our findings underscore the need to rethink aggregation strategies in the face of position bias, paving the way for future research into context-aware methods, such as position-aware voting or debiased sampling, that enhance reliability in long-context tasks.

\section{Related Work}
\label{sec:RelatedWork}

\paragraph{Self-consistency.}
Building on chain-of-thought prompting \citep{wei2022chain}, self-consistency (SC)~\citealp{wang2023selfconsistency}) leverages the intuition that diverse reasoning paths converge toward correct answers when aggregated through methods like majority voting, likelihood scoring \citep{wang2024soft}, LLM Judges \citep{chen2024universal, lin2024just}, or game theoretic consensus \citep{jacob2024consensus}. However, these studies predominantly focus on short tasks (e.g., math problems, brief QA), where stochastic sampling produces \textit{independent errors}. This assumption breaks down in long settings where \textit{systemic errors} emerge due to position bias. While \citet{chen2024universal} evaluates SC on long-context summarization, their reliance on ROUGE metrics, which measure surface-level alignment rather than holistic quality, leaves SC's impact on long-context understanding insufficiently addressed. Our work bridges this gap by comprehensively evaluating SC implementations across multiple long-context tasks, employing automated metrics (ROUGE, F1, accuracy) and LLM judges, with bootstrapped 95\% confidence intervals to rigorously quantify statistical significance.

\paragraph{Position bias.}
Position bias, the preference for certain context regions, is a fundamental challenge for LLMs in long-context processing. Prior work highlights three manifestations: \textit{primacy} (overweighting early context) \citep{wang2023primacy}, \textit{recency} (preferring recent information) \citep{zheng2023large}, and a \textit{U-shaped} pattern where middle-context information is neglected \citep{liu2024lost}. These biases persist across architectures, unaffected by instruction tuning \citep{liu2024lost} or context scaling \citep{lee2024can}, suggesting they are intrinsic to transformer-based attention mechanisms. While recent efforts propose mitigations through context compression \citep{jiang2024longllmlingua}, attention calibration \citep{hsieh2024found}, or fine-tuning \citep{xiong2025from}, such approaches incur computational overhead and merely alleviate symptoms rather than address the root cause. Critically, these approaches target individual model outputs but do not consider how aggregation methods, such as SC, might amplify positional errors—\emph{a gap our work uncovers}. By rigorously quantifying this interaction, we expose SC’s incompatibility with long-context processing and underscore the need for bias-aware aggregation.

\paragraph{Long-context evaluation.}
Long-context evaluation has evolved from early synthetic tasks to more comprehensive benchmarks. \citet{tay2021long} introduced the Long-Range Arena (LRA) to compare efficient transformers, albeit via relatively constrained tasks. More recent efforts, such as RULER \citep{hsieh2024ruler}, extend beyond ``needle-in-a-haystack'' \citep{kamradt2023needle} retrieval by incorporating multi-hop reasoning and aggregation-style tasks. However, \citet{yen2024helmet} raises concerns that synthetic benchmarks may not reliably predict downstream performance, underscoring the need for more representative long-form evaluations. Meanwhile, benchmarks like SCROLLS~\citep{shaham2022scrolls}, its zero-shot variant ZeroSCROLLS~\citep{shaham2023zeroscrolls}, and HELMET~\citep{yen2024helmet} collate tasks featuring more realistic, naturally long text. Our evaluation combines \textit{controlled position probing} and \textit{naturally long tasks}, ensuring that our findings generalize to synthetic and real-world scenarios where systemic position biases persist, regardless of context origin.

\section{Methodology}
\label{sec:Methods}
To systematically evaluate self-consistency (SC) in long-context settings, we design experiments spanning diverse models, task types, and aggregation strategies. Our methodology combines comprehensive benchmarking across real-world tasks with controlled experiments investigating position bias.

\begin{table*}[t]
    \centering
    \small
    {\setlength{\tabcolsep}{5pt}
    \begin{tabular}{lccccc}
        \toprule
        \textbf{Dataset} & \textbf{Split/Source} & \textbf{\#Eval} & \textbf{Task Type} & \textbf{Metric} & \textbf{Avg. \# Words} \\
        \midrule
        GovReport       & LongBench \citep{bai2024longbench} & 200   & Summ          & ROUGE / Judge & 8,734 \\
        QMSum           & LongBench \citep{bai2024longbench} & 200   & QB-Summ       & ROUGE / Judge & 10,614 \\
        SQuALITY        & Test Split \citep{wang2022squality}        & 1040  & QB-Summ       & ROUGE / Judge & 5,208 \\
        Qasper          & LongBench \citep{bai2024longbench} & 200   & QA            & F1            & 3,619 \\
        NarrativeQA     & LongBench \citep{bai2024longbench} & 200   & QA            & F1            & 18,409 \\
        MuSiQue         & LongBench \citep{bai2024longbench} & 200   & Multi-Hop QA  & F1            & 11,214 \\
        QuALITY         & Dev Split \citep{pang2022quality}         & 2086  & MCQA          & Accuracy      & 5,183 \\
        NQ-Open (QA)    & \citet{liu2024lost}                       & 2655  & MDQA          & Accuracy      & - \\
        NQ-Open (TR)    & \citet{liu2024lost}                       & 2655  & Retrieval     & Accuracy      & - \\
        \bottomrule
    \end{tabular}
    }
    \caption{
        \textbf{Overview of tasks and datasets used in our experiments.} 
        The QuALITY dev split was used as the annotations for the test split are not publicly released.
        QB-Summ indicates query-based summarization, MCQA indicates multiple choice question answering, and MDQA indicates multi-document question answering.
        The ``\#Eval'' column refers to the exact number of examples in each dataset split we evaluate. 
        For summarization tasks, we report both ROUGE and a GPT-4o-based Judge score; 
        for QA tasks, we report F1 or accuracy depending on the output format.
        The average number of words in NQ-Open examples varies depending upon the total number of documents used.
    }
    \label{tab:datasets}
\end{table*}

\subsection{Tasks and Datasets}
For our dataset-level evaluations (\S\ref{sec:HESC}), we utilize seven datasets with distinct challenges in long-context processing, from dense information synthesis to targeted retrieval and reasoning (Table~\ref{tab:datasets}). Summarization tasks include GovReport~\citep{huang2021efficient}, QMSum~\citep{zhong2021qmsum}, and SQuALITY~\citep{wang2022squality}, which test dense information synthesis. We use Qasper~\citep{dasigi2021dataset}, NarrativeQA~\citep{kovcisky2018narrativeqa}, and QuALITY~\citep{pang2022quality} for question answering, where models must understand details in long narratives. We also evaluate MuSiQue~\citep{trivedi2022musique} for multi-hop reasoning across documents. Datasets are sourced from LongBench \citep{bai2024longbench}, a standardized benchmark for long-context evaluation, except SQuALITY \citep{wang2022squality} and QuALITY \citep{pang2022quality}, where we use the original test and dev splits, as they were omitted from LongBench.

For our position bias investigation (\S\ref{sec:Phenomenon}), we utilize NQ-Open \citep{kwiatkowski2019natural, lee2019latent}, comprising 2,655 real user queries paired with paragraph-length answers from Wikipedia. We design two tasks:
(1) \underline{Question Answering (QA):} Given a query and a collection of documents (with one gold document), models must generate the correct answer. This task examines the model’s ability to locate and use information from lengthy contexts.
(2) \underline{Text Retrieval (TR):} Models are asked to identify which document contains the correct answer. By isolating the retrieval step, this task enables us to disentangle the effects of self-consistency on information localization from those of answer generation.
These tasks have been used in prior works \citep{liu2024lost, lee2024can} as two canonical long-text tasks, providing a solid foundation for comparing our results with existing literature. Task prompt templates are illustrated in \S\ref{app:TaskTypes}.

\subsection{Self-Consistency Implementations}
To assess self-consistency across diverse task formats, we differentiate our SC implementation based on the nature of the task output. For tasks requiring open-ended generation (GovReport, QMSum, SQUALITY, Qasper, NarrativeQA, MuSiQue, and NQ-Open QA/TR), we primarily employ Universal Self-Consistency (USC), which utilizes an LLM judge for response aggregation. For tasks with discrete, categorical answer options, specifically the multiple-choice question answering dataset QUALITY, we use traditional Majority Voting SC. We also evaluate Soft-SC across a subset of tasks as an alternative aggregation mechanism. By comparing these aggregation strategies across diverse tasks, we ensure our conclusions are not specific to any single SC method. All implementations maintain consistent sampling parameters (eight samples, temperature 1.0) unless explicitly varied for experimental purposes.

\paragraph{SC (for QuALITY).}
We implement traditional majority voting SC as described in \citet{wang2023selfconsistency} for QuALITY's multiple-choice format. This implementation selects the most frequently generated answer across eight samples, breaking ties by selecting the first most frequent answer.

\paragraph{USC (for open-ended tasks).}
Our primary implementation follows the USC baseline of \citet{chen2024universal}, generating eight samples per input at temperature $1.0$. These parameters balance sampling diversity against computational cost while maintaining output quality. For aggregation, we employ GPT-4o as a judge (\S\ref{subsec:USCPrompts}) to select the highest-quality response from the generated candidates, using carefully designed evaluation criteria specific to each task type.

\paragraph{Soft-SC.}
We also implement Soft-SC following \citet{wang2024soft}. This variant replaces discrete voting with a continuous scoring mechanism based on model likelihood. For each generated response, we compute the mean token likelihood across all response tokens. The final output is the response with the highest mean token likelihood, providing a more nuanced selection mechanism incorporating model uncertainty. By replacing discontinuous majority voting with likelihood-based aggregation, Soft-SC tests whether model confidence (rather than frequency) can mitigate correlated errors.

In short, we use USC for open-ended tasks, majority-vote SC for the QuALITY multiple-choice dataset, and soft-SC as a robustness ablation against aggregation mechanism.

\subsection{Model Selection and Configuration}
We evaluate eight instruction-tuned models spanning proprietary and open-source families, ranging from 3B to 72B parameters. All models are deployed in a standardized environment to eliminate implementation-specific confounders. Proprietary models representing state-of-the-art commercial systems include OpenAI's GPT-4o and GPT-4o-mini \citep{openai2024gpt}. For open-source architectures, we test Meta's LLaMA-3-Instruct series \citep[3.2-3B, 3.1-8B, 3.3-70B,][]{grattafiori2024llama} and Alibaba's Qwen-2.5 series \citep[3B, 7B, 72B,][]{yang2024qwen2}. We excluded base models from our analysis as their documented underperformance on generative tasks would confound our investigation of SC's impact. Deployment was via the vLLM framework \citep{kwon2023efficient}, which provides efficient serving with consistent runtime characteristics across all models. Serving configurations and hardware environments (NVIDIA A100 80GB and A6000 48GB) are identical for all experiments.

\subsection{Evaluation Framework}
We employ task-specific performance metrics (ROUGE, F1, accuracy) alongside LLM-based evaluation for summarization to ensure a rigorous assessment of SC's impact. We further quantify performance shifts using bootstrapped 95\% confidence intervals, allowing us to determine whether SC effects are statistically significant.

\paragraph{Multi-metric assessment.}
For summarization, we extend \citet{shaham2023zeroscrolls}'s evaluation protocol, reporting both the geometric mean of ROUGE-1, ROUGE-2, and ROUGE-L \citep{lin2004rouge} and GPT-4o-as-a-judge scores (0–100 scale (\S\ref{subsec:JudgePrompts})). While ROUGE quantifies surface-level alignment, it fails to capture essential quality dimensions, such as factual consistency and narrative flow \citep{fabbri2021summeval}. Our GPT-4o judge evaluates summaries on consistency (alignment with references), relevance (focus on key points), fluency (grammatical coherence), and informativeness (coverage of critical details). SQuALITY additionally uses QA-specific criteria, wherein the judge evaluates summaries on correctness (factual accuracy), relevance (focus on key points), and fluency (grammatical coherence).

For QA tasks (Qasper, NarrativeQA, MuSiQue), we compute unigram F1 scores between outputs and annotated answers to accommodate paraphrasing better than exact match. QuALITY uses accuracy for its closed-set multiple-choice format. For the position bias experiments, we also use accuracy as our primary metric: for QA, accuracy measures the exact match between the generated answer and gold answer, and for TR, accuracy measures whether models correctly identify the source document containing the answer.

\paragraph{Performance difference and statistical rigor.}
To quantify the impact of self-consistency, we measure the direct performance difference:
\begin{equation} \label{eq:sc_difference}
    \text{Difference} = \text{Perf. w/ SC} - \text{Perf. w/o SC}
\end{equation}
Statistical significance of the performance differences was assessed via bootstrapped 95\% confidence intervals using 10K resamples. We use bootstrapped confidence intervals (CIs) to avoid challenges related to the unknowns of the underlying distribution of differences. A CI containing 0 indicates no significant change, while an entirely positive or negative CI indicates a significant performance gain or degradation, respectively.

\paragraph{Summary.}
By employing a multi-metric approach, we ensure that our findings are robust across different evaluation criteria, thereby mitigating the impact of any single metric's limitations. Our analysis presents baseline performance using greedy decoding, compares it to SC performance, and demonstrates the significance of observed differences through bootstrapped 95\% confidence intervals. This evaluation framework ensures our findings are statistically robust while accounting for variance in model outputs and task difficulty.

\section{How Effective is Self-Consistency for Long-Context Problems?}
\label{sec:HESC}
Our experiments reveal systematic challenges for self-consistency (SC) in long-context tasks. In the section that follows, we first analyze performance degradation across tasks (\S\ref{subsec:TaskDgradation}), then examine model-scale trends (\S\ref{subsec:ScaleDependency}), and finally explore how robust these effects are to variations in SC implementations (\S\ref{subsec:RobustToImplementation}).

\subsection{Performance Degradation Across Tasks}
\label{subsec:TaskDgradation}

\begin{table*}[p!]
    \small
    \centering
    \renewcommand{\arraystretch}{1.2}
    \resizebox{.87\textwidth}{!}{
    \begin{tabular}{c|l|cc|cc}
        \hline
        \multirow{2}{*}{\textbf{Dataset}}
        & \multirow{2}{*}{\textbf{Model}}
        & \multicolumn{2}{c|}{\textbf{ROUGE}}
        & \multicolumn{2}{c}{\textbf{Judge}} \\
        \cline{3-6}
        &
        & \textbf{Difference} & \textbf{95\% CI}
        & \textbf{Difference} & \textbf{95\% CI} \\
        \hline
        \multirow{8}{*}{\rotatebox{90}{GovReport (USC)}}
        & GPT-4o
          & $-2.2_{(23.6\leftarrow25.8)}$ & {\color{red}[-2.6, -1.8]}
          & $-0.8_{(80.0\leftarrow80.8)}$ & [-2.3, 0.6] \\
        & GPT-4o-Mini
          & $-1.7_{(23.8\leftarrow25.5)}$ & {\color{red}[-2.0, -1.4]}
          & $0.5_{(79.1\leftarrow78.6)}$ & [-1.4, 2.2] \\
        & LLaMA-3.3-70B
          & $0.1_{(29.5\leftarrow29.4)}$  & [-0.4, 0.5]
          & $-0.6_{(81.6\leftarrow82.2)}$ & [-2.1, 0.8] \\
        & LLaMA-3.1-8B
          & $-1.1_{(29.4\leftarrow30.5)}$ & {\color{red}[-1.5, -0.6]}
          & $-0.1_{(80.1\leftarrow80.2)}$ & [-1.9, 1.5] \\
        & LLaMA-3.2-3B 
          & $-2.2_{(27.0\leftarrow29.2)}$ & {\color{red}[-2.8, -1.6]}
          & $-1.0_{(76.8\leftarrow77.8)}$ & [-3.3, 0.9] \\
        & Qwen-2.5-72B
          & $-0.7_{(28.4\leftarrow29.1)}$ & {\color{red}[-1.1, -0.3]}
          & $0.3_{(80.7\leftarrow80.4)}$  & [-1.4, 1.9] \\
        & Qwen-2.5-7B 
          & $-2.7_{(25.8\leftarrow28.7)}$ & {\color{red}[-3.3, -2.4]}
          & $-0.9_{(79.8\leftarrow80.7)}$ & [-2.1, 0.4] \\
        & Qwen-2.5-3B
          & $-5.0_{(23.9\leftarrow28.9)}$ & {\color{red}[-5.6, -4.5]}
          & $-2.3_{(76.4\leftarrow78.7)}$ & {\color{red}[-4.4, -0.3]} \\
        \hline
        \multirow{8}{*}{\rotatebox{90}{QMSum (USC)}}
        & GPT-4o
          & $-1.0_{(17.8\leftarrow18.8)}$ & {\color{red}[-1.5, -0.5]}
          & $-0.2_{(58.0\leftarrow58.2)}$ & [-2.6, 2.1] \\
        & GPT-4o-Mini
          & $-0.3_{(18.0\leftarrow18.3)}$ & [-0.9, 0.3]
          & $-0.7_{(57.9\leftarrow58.6)}$ & [-3.1, 1.5] \\
        & LLaMA-3.3-70B
          & $-0.4_{(18.9\leftarrow19.3)}$ & [-1.1, 0.3]
          & $0.5_{(57.6\leftarrow57.1)}$  & [-2.0, 3.1] \\
        & LLaMA-3.1-8B
          & $-1.3_{(17.4\leftarrow18.7)}$ & {\color{red}[-2.0, -0.5]}
          & $0.4_{(53.6\leftarrow53.2)}$   & [-1.9, 2.8] \\
        & LLaMA-3.2-3B
          & $-1.3_{(15.3\leftarrow16.6)}$ & {\color{red}[-2.1, -0.6]}
          & $0.7_{(48.8\leftarrow48.1)}$  & [-2.0, 3.3] \\
        & Qwen-2.5-72B
          & $-0.6_{(17.8\leftarrow18.4)}$ & [-1.1, 0.1]
          & $-0.4_{(56.8\leftarrow57.2)}$ & [-2.7, 2.0] \\
        & Qwen-2.5-7B
          & $-0.8_{(17.4\leftarrow18.6)}$ & {\color{red}[-1.8, -0.7]}
          & $-0.4_{(56.4\leftarrow56.8)}$ & [-2.7, 2.0] \\
        & Qwen-2.5-3B 
          & $-2.3_{(14.4\leftarrow16.7)}$ & {\color{red}[-3.0, -1.7]}
          & $-1.5_{(50.8\leftarrow52.3)}$ & [-4.0, 1.2] \\
        \hline
        \multirow{8}{*}{\rotatebox{90}{SQuALITY (USC)}}
        & GPT-4o
          & $-1.5_{(17.2\leftarrow18.7)}$ & {\color{red}[-1.7, -1.3]}
          & $-1.1_{(80.9\leftarrow82.0)}$ & {\color{red}[-1.9, -0.4]} \\
        & GPT-4o-Mini
          & $-1.0_{(16.4\leftarrow17.4)}$ & {\color{red}[-1.1, -0.8]}
          & $-0.4_{(79.5\leftarrow79.9)}$ & [-1.2, 0.4] \\
        & LLaMA-3.3-70B
          & $-0.1_{(19.6\leftarrow19.7)}$ & [-0.2, 0.1]
          & $0.5_{(80.1\leftarrow79.6)}$  & [-0.3, 1.2] \\
        & LLaMA-3.1-8B
          & $-1.2_{(18.3\leftarrow19.5)}$ & {\color{red}[-1.4, -1.0]}
          & $0.1_{(77.4\leftarrow77.3)}$  & [-0.7, 0.9] \\
        & LLaMA-3.2-3B
          & $-1.7_{(16.7\leftarrow18.4)}$ & {\color{red}[-2.0, -1.5]}
          & $-0.3_{(75.6\leftarrow75.9)}$ & [-1.1, 0.5] \\
        & Qwen-2.5-72B
          & $-0.6_{(18.9\leftarrow19.5)}$ & {\color{red}[-0.7, -0.4]}
          & $-1.6_{(79.9\leftarrow81.6)}$ & {\color{red}[-2.4, -0.9]} \\
        & Qwen-2.5-7B
          & $-1.3_{(17.6\leftarrow18.9)}$ & {\color{red}[-1.4, -1.1]}
          & $-1.3_{(77.2\leftarrow78.5)}$ & {\color{red}[-2.1, -0.6]} \\
        & Qwen-2.5-3B
          & $-2.6_{(16.1\leftarrow18.7)}$ & {\color{red}[-2.8, -2.4]}
          & $-1.5_{(75.5\leftarrow77.0)}$ & {\color{red}[-2.3, -0.8]} \\
    \end{tabular}
    }
    \resizebox{.99\textwidth}{!}{
    \begin{tabular}{c|@{\hspace{0.5em}}l@{\hspace{0.5em}}c@{\hspace{0.5em}}c}
        \hline
        \textbf{Dataset} & \textbf{Model} & \textbf{Difference} & \textbf{95\% CI} \\
        \hline
        \multirow{8}{*}{\rotatebox{90}{Qasper (USC)}}
        & GPT-4o & $3.3_{(51.5\leftarrow48.2)}$ & \textcolor{ForestGreen}{[1.0, 5.9]} \\
        & GPT-4o-Mini & $0.1_{(48.0\leftarrow47.9)}$ & [-1.9, 2.0] \\
        & LLaMA-3.3-70B & $-2.5_{(49.4\leftarrow51.9)}$ & \textcolor{red}{[-5.0, -0.4]} \\
        & LLaMA-3.1-8B & $-15.1_{(32.0\leftarrow47.1)}$ & \textcolor{red}{[-19.6, -11.0]} \\
        & LLaMA-3.2-3B & $-1.3_{(38.9\leftarrow40.2)}$ & [-4.9, 2.3] \\
        & Qwen-2.5-72B & $-0.8_{(49.7\leftarrow50.5)}$ & [-3.0, 1.0] \\
        & Qwen-2.5-7B & $-7.3_{(41.1\leftarrow48.4)}$ & \textcolor{red}{[-11.9, -3.4]} \\
        & Qwen-2.5-3B & $-10.8_{(31.3\leftarrow42.1)}$ & \textcolor{red}{[-15.9, -6.5]} \\
        \hline
        \multirow{8}{*}{\rotatebox{90}{Narrative QA (USC)}}
        & GPT-4o & $0.3_{(30.1\leftarrow29.8)}$ & [-1.6, 1.6] \\
        & GPT-4o-Mini & $0.6_{(28.2\leftarrow27.6)}$ & [-0.4, 2.0] \\
        & LLaMA-3.3-70B & $-0.7_{(15.8\leftarrow16.5)}$ & [-2.2, 0.8] \\
        & LLaMA-3.1-8B & $-2.3_{(10.6\leftarrow12.9)}$ & \textcolor{red}{[-4.7, -0.6]} \\
        & LLaMA-3.2-3B & $-0.6_{(8.2\leftarrow8.8)}$ & [-2.6, 1.2] \\
        & Qwen-2.5-72B & $0.4_{(15.5\leftarrow15.1)}$ & [-0.6, 1.7] \\
        & Qwen-2.5-7B & $-0.2_{(11.7\leftarrow11.9)}$ & [-2.3, 2.1] \\
        & Qwen-2.5-3B & $-1.7_{(9.5\leftarrow11.2)}$ & [-4.6, 0.2] \\
        \hline
    \end{tabular}
    \begin{tabular}{|c|@{\hspace{0.5em}}l@{\hspace{0.5em}}c@{\hspace{0.5em}}c}
        \hline
        \textbf{Dataset} & \textbf{Model} & \textbf{Difference} & \textbf{95\% CI} \\
        \hline
        \multirow{8}{*}{\rotatebox{90}{MuSiQue (USC)}}
        & GPT-4o & $0.3_{(38.5\leftarrow38.2)}$ & [-2.6, 3.4] \\
        & GPT-4o-Mini & $2.8_{(36.1\leftarrow33.3)}$ & \textcolor{ForestGreen}{[0.2, 5.8]} \\
        & LLaMA-3.3-70B & $2.5_{(44.7\leftarrow42.2)}$ & [-0.5, 5.6] \\
        & LLaMA-3.1-8B & $4.6_{(25.5\leftarrow20.9)}$ & \textcolor{ForestGreen}{[1.7, 8.4]} \\
        & LLaMA-3.2-3B & $-3.0_{(17.9\leftarrow20.9)}$ & [-6.6, 0.1] \\
        & Qwen-2.5-72B & $-0.2_{(50.3\leftarrow50.5)}$ & [-2.6, 1.8] \\
        & Qwen-2.5-7B & $0.9_{(29.6\leftarrow28.7)}$ & [-2.3, 4.1] \\
        & Qwen-2.5-3B & $0.3_{(12.8\leftarrow12.5)}$ & [-3.0, 3.5] \\
        \hline
        \multirow{8}{*}{\rotatebox{90}{QuALITY (Maj. Vote)}}
        & GPT-4o & $-1.0_{(90.4\leftarrow91.4)}$ & \textcolor{red}{[-1.7, -0.3]} \\
        & GPT-4o-Mini & $-0.3_{(79.7\leftarrow80.0)}$ & [-1.1, 0.4] \\
        & LLaMA-3.3-70B & $-0.3_{(87.0\leftarrow87.3)}$ & [-0.8, 0.1] \\
        & LLaMA-3.1-8B & $-6.0_{(63.7\leftarrow69.7)}$ & \textcolor{red}{[-7.8, -4.3]} \\
        & LLaMA-3.2-3B & $-0.7_{(52.4\leftarrow53.1)}$ & [-2.4, 1.0] \\
        & Qwen-2.5-72B & $-0.7_{(86.8\leftarrow87.5)}$ & \textcolor{red}{[-1.3, -0.1]} \\
        & Qwen-2.5-7B & $0.0_{(74.1\leftarrow74.1)}$ & [-0.6, 0.7] \\
        & Qwen-2.5-3B & $-0.3_{(63.6\leftarrow63.9)}$ & [-0.9, -0.4] \\
        \hline
    \end{tabular}
    }
    \caption{SC and USC performance difference across tasks and models.
    Universal Self-Consistency (USC) is used for all tasks, except QuALITY, which uses majority voting, for aggregation.
    \textit{Red} indicate statistically significant degradation (95\% CI); \textit{green} highlights rare improvements.
    \textbf{Application of SC or USC leads to virtually no significant gains across many long-context task-model pairs (53 of 56 pairs show no statistically significant improvement).} 
    Larger models (LLaMA-70B, Qwen-72B) show milder degradation but no consistent gains, underscoring self-consistency's fundamental limitations.
    }
    \label{tab:main}
\end{table*}

\begin{figure*}[t!]
    \centering
    \begin{subfigure}{0.48\linewidth}
        \centering
        \caption{Qwen models}
        \includegraphics[width=\linewidth,trim=0cm 0.3cm 0cm 2.3cm, clip=true]{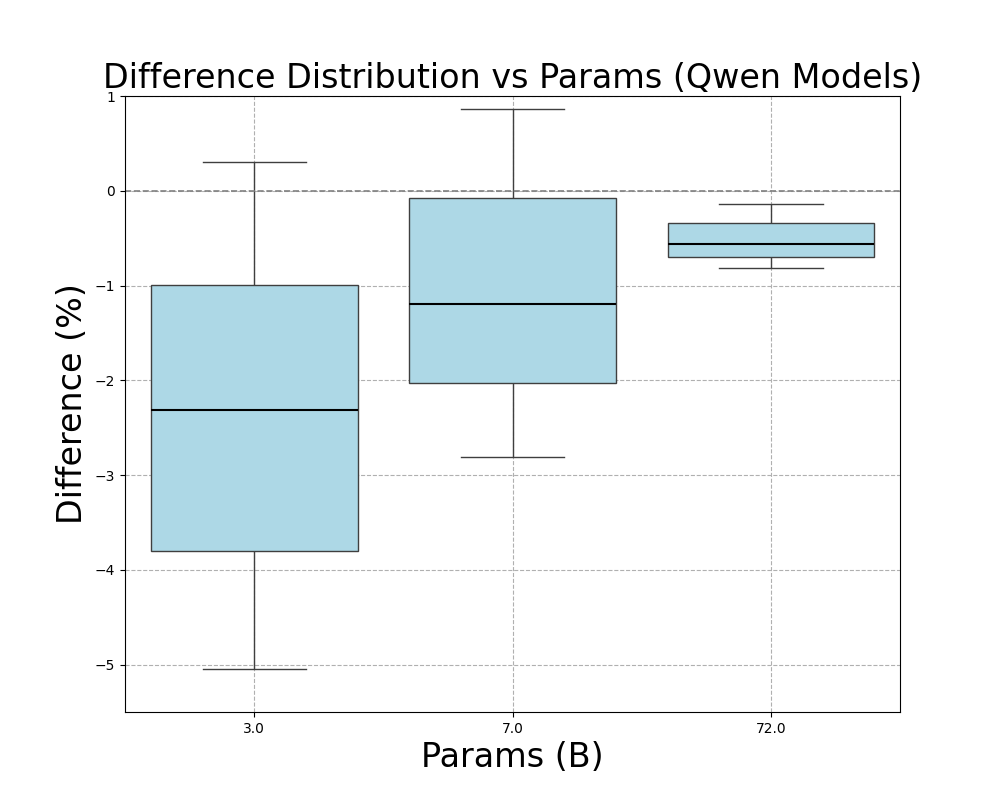}
        
    \end{subfigure}
    \hfill
    \begin{subfigure}{0.48\linewidth}
        \centering
        \caption{LLaMA models}
        \includegraphics[width=\linewidth,trim=0cm 0.3cm 0cm 2.3cm, clip=true]{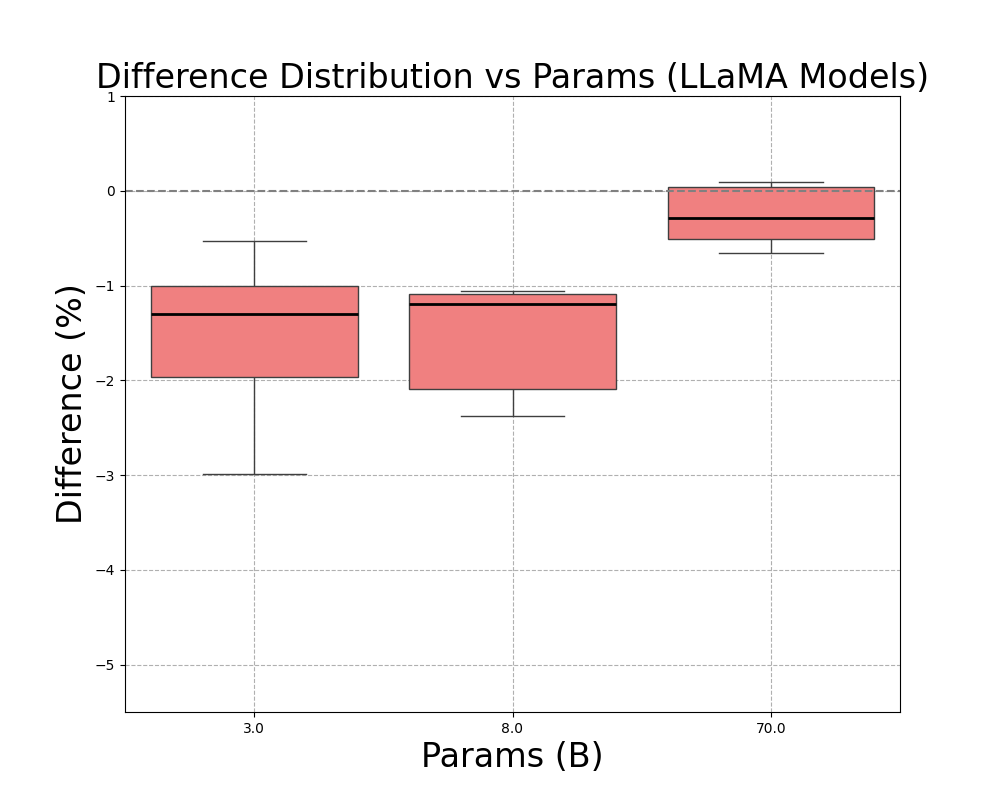}
    \end{subfigure}
    \caption{Average performance difference distribution for Qwen (left) and LLaMA (right) models across all tasks (excluding NQ-Open) in Table~\ref{tab:main}. The y-axis shows the difference in performance between SC and baseline approaches (negative values indicate degradation). Box plots show quartiles with whiskers extending to min/max values. Both model families demonstrate reduced performance degradation as model size increases, \textbf{but even the largest models still fail to break-even}.}
    \label{fig:difference_box}
\end{figure*}

Table~\ref{tab:main} shows the performance difference when applying self-consistency across numerous tasks and models, along with the specific aggregation method used for each task (USC for open-ended generation, Majority Voting for QuALITY). \textbf{Applying majority voting SC or USC leads to virtually no significant gains across many long-context task-model pairs.} Among 56 dataset-model pairs, only three illustrate statistically significant improvement from applying SC or USC.

ROUGE metrics demonstrate significant declines in 79\% of model-dataset pairs (19/24), with smaller models experiencing the most severe degradation (e.g., Qwen-3B: $-5.0$ ROUGE). While LLM Judge scores show more modest effects, they still indicate degradation in 21\% of cases (5/24), with no instances of significant improvement. This consistent pattern across surface-level alignment and holistic judgment metrics suggests that SC fundamentally compromises summarization quality.

While results show slightly more variation in the QA domain, they still predominantly challenge SC's effectiveness. Of 32 model-dataset pairs, only three exhibited statistically significant improvements: GPT-4o on Qasper ($3.3_{(51.5\leftarrow48.2)}$), GPT-4o-mini on MuSiQue ($2.8_{(36.1\leftarrow33.3)}$), and LLaMA-3.1-8B on MuSiQue ($4.6_{(25.5\leftarrow20.9)}$). However, these improvements are isolated cases rather than indicative of a broader trend. Notably, LLaMA-3.1-8B's performance varies dramatically across datasets, showing improvement on MuSiQue but suffering a severe $-15.1_{(32.0\leftarrow47.1)}$ degradation on Qasper, which we attribute to poor instruction following despite adequate performance on other datasets. Eight pairs show significant degradation, particularly among smaller models.

\paragraph{Robustness to choice of metric.}
Despite differences in magnitude, the consistency between ROUGE and LLM Judge metrics in indicating performance degradation suggests our findings reflect genuine limitations of SC rather than artifacts of specific evaluation methods. The more moderate effects observed in LLM Judge scores indicate that human-like evaluation demonstrates greater robustness to surface-level variations in outputs while still capturing the overall negative impact of SC on model performance.

\subsection{Model-Scale Dependency}
\label{subsec:ScaleDependency}
Figure~\ref{fig:difference_box} shows scaling between model size and performance degradation under SC and USC, with greater declines in smaller models. We compute Pearson and Spearman correlations between model size and performance difference. We find weak-to-moderate monotonicity, with stronger effects under ROUGE (Spearman’s $\rho=0.50$, $p<0.01$) than LLM-Judge (Spearman’s $\rho=0.31$, $p<0.05$). However, this relationship is nonlinear, as the Pearson correlation is \emph{not} statistically significant under either ROUGE nor the LLM Judge. While larger models exhibit more stability than smaller ones, this ``resistance'' to degradation rarely translates into actual improvements. This pattern suggests that \textbf{while increased model scale might mitigate SC's adverse effects, it fails to unlock SC's promised benefits in long-context scenarios.}

\subsection{Robustness to SC Aggregation Method}
\label{subsec:RobustToImplementation}
Having established the generally negative impact of USC (on generative tasks) and Majority Voting SC (on QUALITY), we further investigated whether an alternative aggregation strategy could alter these findings. To ensure our findings are not artifacts of our primary SC implementations, we evaluated Soft SC on a subset of tasks (Table~\ref{tab:softsc}).

\begin{table}[ht!]
    \small
    \centering
    \begin{tabular}{l@{\hspace{0.7em}}c@{\hspace{0.7em}}c}
        \hline
        \textbf{Model} & \textbf{Difference} & \textbf{95\% CI} \\
        \hline
        \multicolumn{3}{l}{GovReport (ROUGE)} \\
        GPT-4o & $-1.8_{(24.0\leftarrow25.8)}$ & \textcolor{red}{[-2.2, -1.4]} \\
        GPT-4o-mini & $-1.3_{(24.2\leftarrow25.5)}$ & \textcolor{red}{[-1.6, -1.0]} \\
        \hline
        \multicolumn{3}{l}{GovReport (Judge)} \\
        GPT-4o & $-1.3_{(79.5\leftarrow80.8)}$ & [-2.6, 0.1] \\
        GPT-4o-mini & $-0.4_{(78.2\leftarrow78.6)}$ & [-2.0, 1.2] \\
        \hline
        \multicolumn{3}{l}{Qasper} \\
        GPT-4o & $-0.1_{(48.1\leftarrow48.2)}$ & [-3.5, 2.8] \\
        GPT-4o-mini & $0.1_{(48.0\leftarrow47.9)}$ & [-1.6, 1.3] \\
        \hline
        \multicolumn{3}{l}{MuSiQue} \\
        GPT-4o & $-1.9_{(36.3\leftarrow38.2)}$ & [-4.9, 0.5] \\
        GPT-4o-mini & $1.5_{(34.8\leftarrow33.3)}$ & [-0.7, 4.0] \\
        \hline
    \end{tabular}
    \caption{Soft self-consistency results on select datasets. Alternative SC implementation fails to show improvement, showing significant degradation (red) on GovReport (ROUGE). Results suggest \textbf{SC's limitations persist across implementation strategies.} Performance patterns mirror standard SC across metrics and model sizes.}
    \label{tab:softsc}
\end{table}

\paragraph{SC's limitation persists across implementation strategies.}
This approach fails to improve over baseline, showing significant degradation on GovReport (ROUGE) and no significant gains across other tasks. The consistent degradation observed with both standard SC / USC and Soft SC implementations suggests that the limitations are fundamental to the SC approach rather than artifacts of specific implementation choices. This consistent pattern of failure across models, tasks, and implementations strengthens our central hypothesis: that position bias undermines SC's core assumption of error independence by inducing strongly correlated errors. This suggests a systematic, mechanistic cause, which we isolate in the following section through a series of controlled experiments designed to explain the observed performance degradations. Together with the SC and USC results in Table~\ref{tab:main}, these findings demonstrate that SC's limitations persist regardless of aggregation strategy.

\section{Understanding the Phenomenon: Position Bias and Self-Consistency}
\label{sec:Phenomenon}
To probe the failures cataloged in \S\ref{sec:HESC}, we ask if \textit{positional bias} creates correlated errors that self-consistency (SC) merely amplifies. We test (i) context-length ablations (\S\ref{subsec:CorrelatedErrors}), (ii) paired QA–retrieval evaluations (\S\ref{subsec:CorrelatedErrors}), and (iii) robustness checks (\S\ref{subsec:Robustness}), each confirming that the degradation stems from position-induced bias, not SC.

\subsection{Controlled Exploration of Position Bias}
We evaluate SC's impact through controlled experiments on question answering (QA) and text retrieval (TR) tasks (\S\ref{sec:Methods}). The QA task tests information synthesis, while TR isolates information retrieval, allowing us to separate position effects on access from those on reasoning. Our experimental design follows that of \citet{liu2024lost}, using their variant of NQ-Open with varying context sizes (10, 20, and 30 documents) and gold document positions (beginning, middle, or end) to assess position sensitivity. We employ three complementary metrics to quantify performance: QA accuracy measures answer correctness, TR accuracy evaluates document identification success, and the performance delta (Equation~\ref{eq:sc_difference}), which reveals SC's relative impact across positions. To ensure our findings are robust to implementation choices, we evaluate multiple configurations by varying sample counts (4, 8, 16) and temperatures (0.2, 1.0, 1.8).

Extensive literature \citep[\emph{inter alia}]{perez2021true, Lu2022FantasticallyOP, mishra2022reframing} has demonstrated that LLM performance is susceptible to the language and structure of their prompts. To examine the effects of prompt formatting, we test three distinct prompt formulations: documents-then-question (Doc-Q), question-then-documents (Q-Doc), and a query-aware approach with the question bracketing the documents (Q-Doc-Q).

\subsection{Evidence of Correlated Errors}
\label{subsec:CorrelatedErrors}
Our analysis reveals that SC's impact varies systematically with document position, fundamentally challenging previous assumptions about its effectiveness. The data demonstrate distinct patterns across task types and positions, providing strong evidence for position-induced correlated errors.

\begin{figure*}[p!] 
    \centering
    \begin{subfigure}[t]{\linewidth}
        \centering
        \caption{Question Answering}
        \resizebox{\width}{0.95\height}{\includegraphics[trim={0cm 1.5cm 0cm 0cm},clip,width=\linewidth]{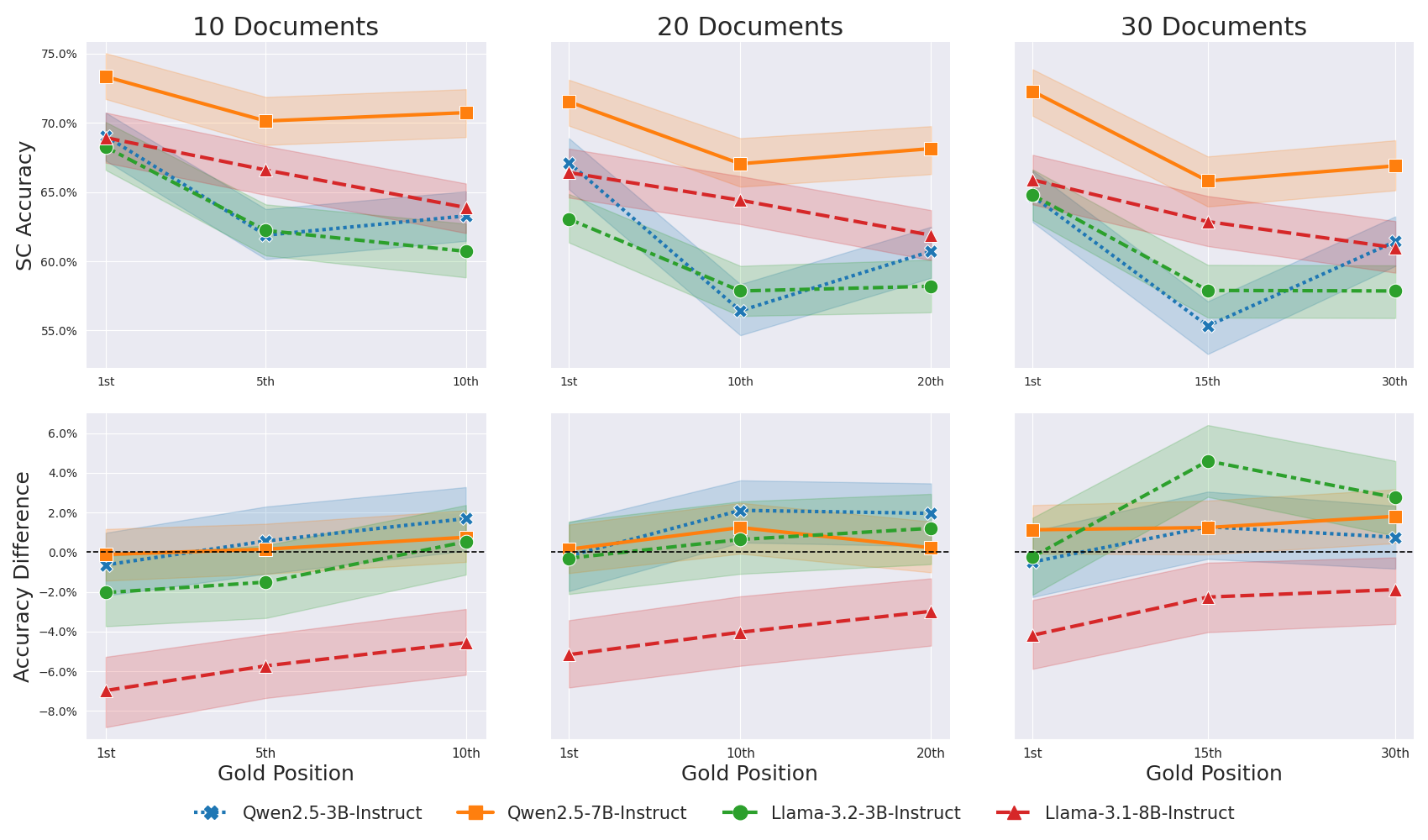}}
        \label{fig:stacked_qa}
    \end{subfigure}

    \vspace{-1.5em}
    
    \begin{subfigure}[t]{\linewidth}
        \centering
        \caption{Text Retrieval}
        \resizebox{\width}{0.95\height}{\includegraphics[trim={0cm 2.3cm 0cm 0cm},width=\linewidth]{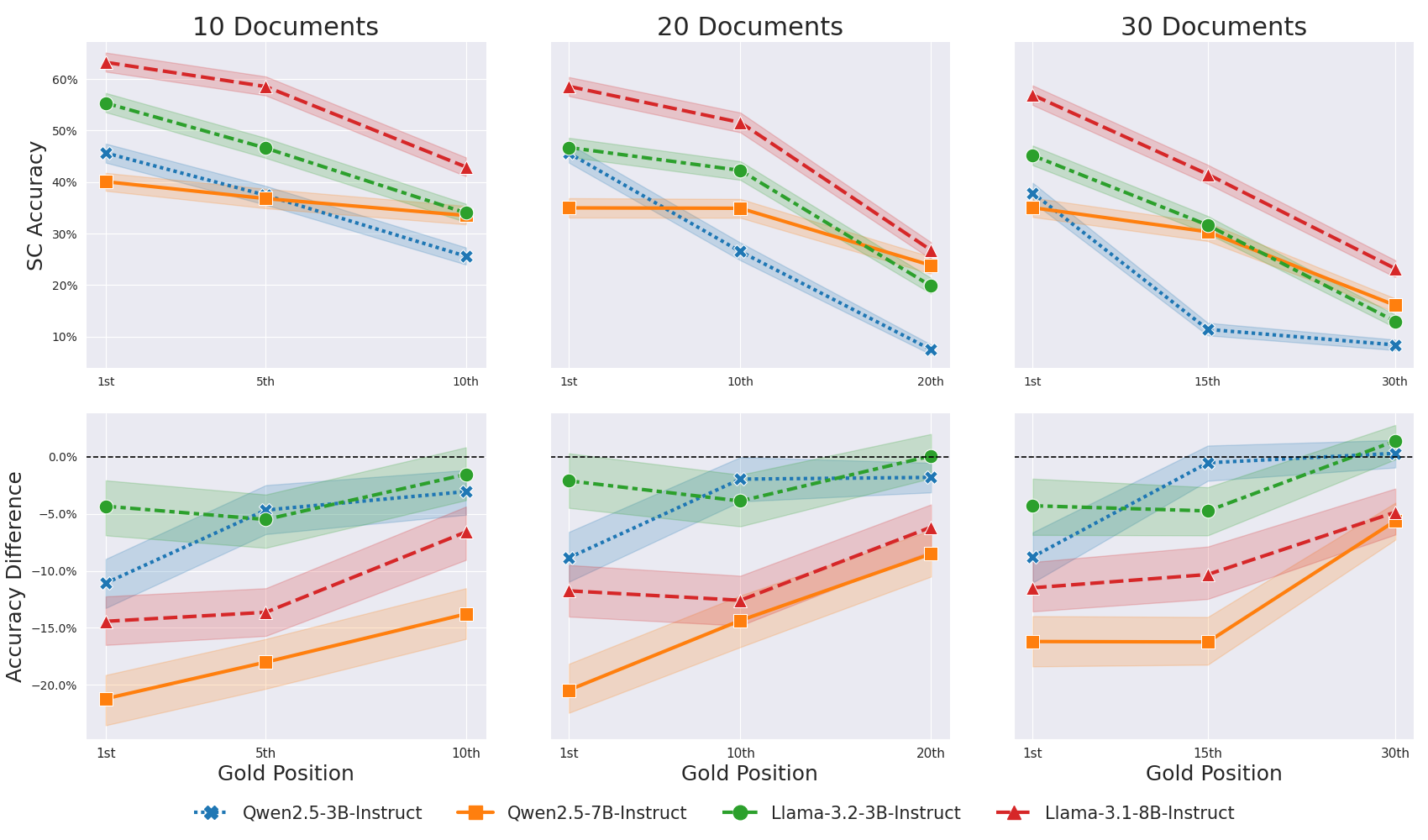}}
        \label{fig:stacked_tr}
    \end{subfigure}

    \caption{Self-consistency accuracy across models for NQ-Open, showing positional bias. (a) QA accuracy is highest at the beginning or end of context, with LLaMA-3.1 degrading under SC across positions; a U-shaped pattern persists across context lengths and model sizes. The accuracy difference reveals an upward trend as the gold document position moves later in context--\textbf{SC provides relatively less harm for later positions but never achieves the performance levels of baseline models on early positions.} (b) TR accuracy also peaks at the start, with severe drops as context length and position of relevant information increase. The corresponding difference plots demonstrate consistent negative impact across positions, with particularly severe degradation (20-25\%) for early positions in longer contexts.}
    \label{fig:stacked_qa_tr}
\end{figure*}

\paragraph{Question answering tasks.}
In QA tasks, SC performance exhibits a distinctive U-shaped curve (Figure~\ref{fig:stacked_qa}), with accuracy peaking when relevant information appears at context boundaries but dropping significantly for middle positions; this pattern persists across all models and context lengths. The position sensitivity becomes more pronounced as context length increases--for instance, in 30-document contexts, Qwen-2.5-7B shows a $15\% \pm 2\%$ accuracy drop for middle positions compared to boundary positions. 

When examining the performance delta, the relationship between position and SC effectiveness becomes more apparent. This reveals a consistent upward trend as the gold document moves later in the context, suggesting that SC's harmful effect is relatively lessened for information appearing at the end of the context. However, this requires careful interpretation: while the performance delta improves for later positions, absolute performance under SC never matches the baseline on early-position documents, demonstrating that \textbf{SC may partially compensate for position-related degradation but cannot overcome the underlying position bias}.

\begin{figure*}[ht]
    \centering
    \resizebox{\width}{0.95\height}{\includegraphics[width=\linewidth]{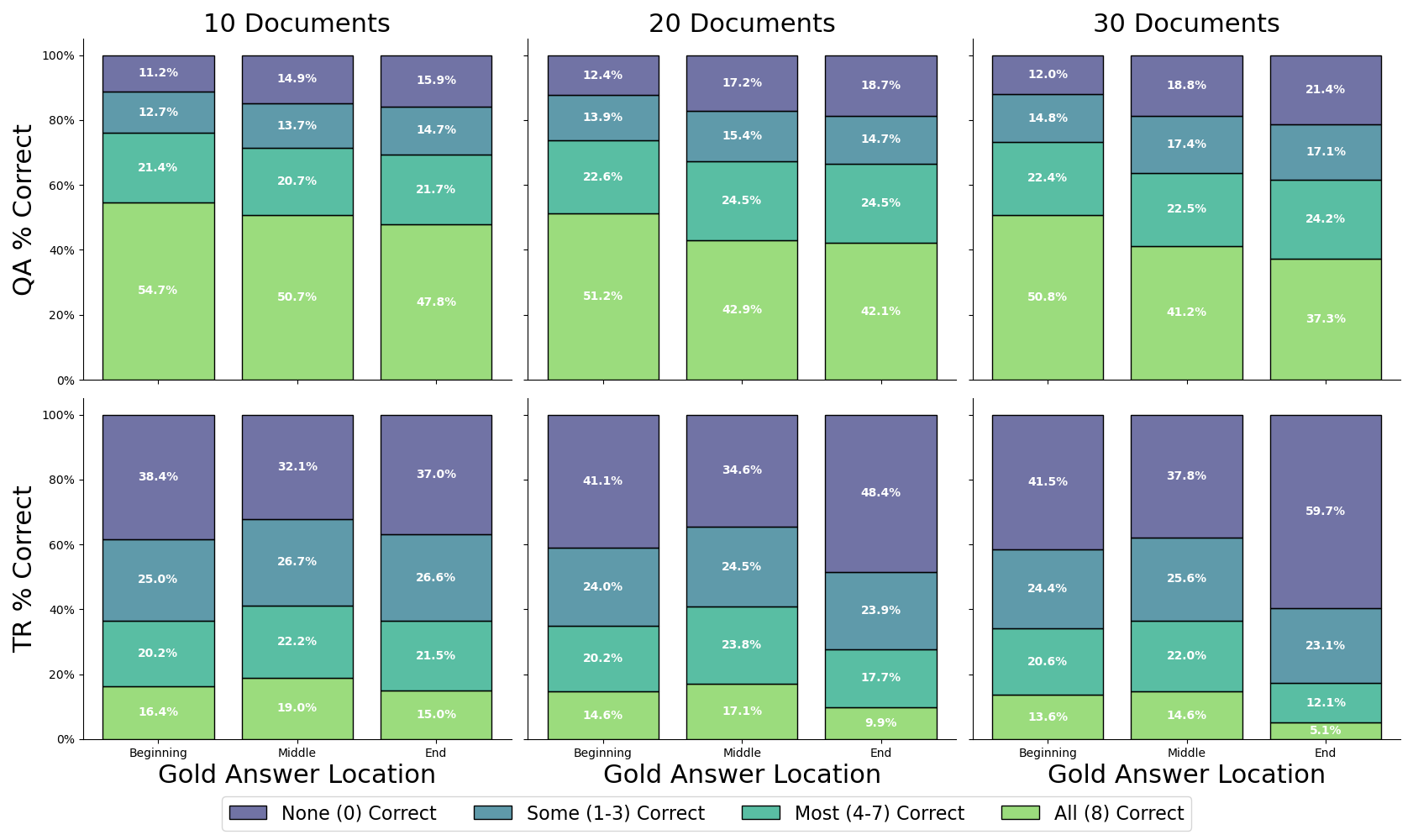}}
    \caption{Distribution of correctness across eight intermediate generations, demonstrating positional bias before self-consistency aggregation. The stacked bars show the percentage of cases where none (0), some (1-3), most (4-7), or all (8) of the intermediate samples were correct. (Top Row) For Question Answering (QA), the prevalence of highly correct sample sets (green bars) follows a U-shaped curve, degrading when information is in the middle. (Bottom Row) For Text Retrieval (TR), correctness declines monotonically, with a dramatic increase in cases where zero samples are correct as the gold document is placed later in longer contexts. This visualization confirms that \textbf{the errors amplified by SC are systemic and correlated}, originating from the model's fundamental bias rather than the aggregation process.}
    \label{fig:stacked_sc_interim_position_bias}
\end{figure*}

\newpage
\paragraph{Text retrieval tasks.}
TR tasks reveal a more severe, qualitatively different manifestation of position bias (Figure~\ref{fig:stacked_tr}). Unlike QA's U-shaped curve, TR accuracy shows a monotonic decline as the gold document moves later in the context. The performance degradation becomes particularly severe with longer contexts—in 30-document scenarios, accuracy drops by $40\% \pm 3\%$ when the gold document appears in the final third of the context. SC consistently deteriorates performance across all positions, with larger models like Qwen-2.5-7B and LLaMA-3.1-8B showing significant 20-25\% performance reductions. To illustrate this failure mode, when the gold document appears early in a 30-document context, Qwen-2.5-7B generates multiple samples that consistently focus on later, irrelevant documents, amplifying its position bias through repeated sampling. The consistency of this pattern provides strong evidence that \textbf{SC not only fails to mitigate position bias but can actively exacerbate it}, particularly in tasks requiring precise information retrieval.

\paragraph{Analysis of intermediate generations.}
To trace the origin of these aggregation-level failures, we analyzed the distribution of correct answers within the intermediate generations before SC is applied (Figure~\ref{fig:stacked_sc_interim_position_bias}). This analysis reveals that the biases observed in the final aggregated output are deeply rooted in the initial sampling step. For QA tasks, the proportion of cases where all or most of the eight samples are correct follows the same U-shaped pattern seen in the final output, dropping when the gold answer is in the middle of the context. The effect is even more pronounced in TR tasks, which exhibit a stark monotonic decline. For instance, in 30-document contexts with the answer at the end, nearly 60\% of the time, none of the eight intermediate samples correctly identified the source document. This provides direct evidence that the \textbf{errors are strongly correlated across samples}, stemming from the model's inherent positional bias and violating the core assumption of independence that underpins self-consistency.

\subsection{Robustness Analysis}
\label{subsec:Robustness}
We conduct extensive robustness checks across prompt structure and sampling parameters, known to sway LLM behavior \citep{perez2021true, Lu2022FantasticallyOP, mishra2022reframing}. Our results demonstrate that implementation choices can affect absolute performance but cannot resolve the underlying challenges of position bias.

\paragraph{Prompt engineering effects.}
Even carefully engineered prompts cannot overcome position-dependent performance degradation. Among the three tested formats, the query-aware approach (Q-Doc-Q) shows a modest advantage (+2-3\%) for early positions in QA tasks. However, this benefit diminishes with longer contexts and middle/end positions (Figure~\ref{fig:qa_prompt}). TR tasks demonstrate even more striking limitations: while format choice can influence overall accuracy by up to 20\%, all formats suffer severe degradation for later positions in long contexts. For example, the documents-then-question format (Doc-Q) drops to approximately $10\% \pm 2\%$ accuracy in 30-document contexts when the target document appears in the final third, regardless of prompt engineering efforts.

\paragraph{Parameter sensitivity.}
Varying SC implementation parameters reveal similar limitations. Increasing sample count yields diminishing returns, where doubling generations from 8 to 16 yields less than 1\% improvement in absolute accuracy while doubling the computational overhead in the QA task. More concerning, larger sample counts actively harm TR performance, likely because they amplify the model's tendency to fixate on positionally favored but incorrect documents. Temperature variations primarily affect overall performance levels rather than positional patterns, with extreme temperatures leading to universal degradation.

\section{Discussion}
\label{sec:Discussion}
Our findings reveal a fundamental limitation in applying self-consistency (SC) to long-context tasks. Whether using sophisticated judge-based aggregation for open-ended generation or traditional majority voting for multiple-choice questions, the result is the same: performance either degrades or fails to improve. This pattern persists across both automated metrics and human-like evaluations, indicating a systemic issue. We argue that this failure stems from positional bias, which induces correlated errors that SC incorrectly amplifies.

\subsection{Effects of Position Bias on Long-Context}
SC was designed to filter uncorrelated errors by leveraging diverse reasoning paths, but our experiments show this assumption fails in long contexts. Positional bias induces strongly correlated errors across samples, and these biases (e.g., over-relying on early context) persist despite varying sample counts or temperature. While one might argue that these failures are not specific to SC but are merely a symptom of models struggling with the difficulty of long-context reasoning, our controlled experiments refute this. The fact that model performance is systematically dependent on the position of the correct information, not just its presence in a long context, demonstrates that the issue is biased information access, which SC incorrectly amplifies.

Building on this insight, we find that SC reinforces these positional blind spots, effectively amplifying rather than mitigating systemic errors. This amplification is a direct result of the aggregation process operating on a set of samples already skewed by bias, where, as shown in our analysis of intermediate generations (Figure~\ref{fig:stacked_sc_interim_position_bias}), a majority of reasoning paths are often predisposed to the same positional error. This mechanism manifests most dramatically in our text retrieval experiments, where SC led to 20-25\% performance reductions in models like Qwen-2.5-7B and LLaMA-3.1-8B.

We frequently observed that a majority of samples incorrectly selected documents from early positions when the correct document was located in the middle or at the end, resulting in a unanimous but incorrect consensus. For instance, in Appendix~\ref{app:ErrorAnalysis}, we illustrate an example in which seven of the eight generations incorrectly identify information from the first document as being correct when, in actuality, the gold document was at the end. These findings suggest that traditional aggregation methods may be fundamentally unsuited for long-context tasks, motivating exploration into position-aware aggregation methods that explicitly account for and counteract positional effects.

\subsection{Practical Implications and Future Directions}
The unsuitability of standard SC for long-context tasks has significant practical implications. In critical fields such as legal analysis or medical diagnostics, overlooking key information buried in a lengthy document can have severe consequences. Practitioners cannot assume that this common inference-time technique will improve reliability and should instead explore alternatives.

Several directions emerge from our findings. First, position-aware aggregation methods could explicitly account for document position when combining multiple samples, perhaps by weighting responses based on their attention patterns across the context. Second, contrastive sampling strategies might reduce the correlation between errors by explicitly encouraging diversity in the context regions to which models attend across samples. Third, attention recalibration techniques, such as those proposed by \citet{hsieh2024found}, could be adapted specifically for sampling-based methods to mitigate positional effects before aggregation. Finally, retrieval-augmented approaches could be combined with SC, allowing models to process small context windows where traditional SC remains effective.

These approaches represent testable hypotheses for follow-up work addressing the challenge of correlated positional errors. Future work may unlock more reliable performance in long-context scenarios while maintaining the benefits of aggregation by developing techniques that directly counter position bias rather than simply applying SC.

\section{Conclusion}
\label{sec:Conclusion}
Our study provides the first in-depth analysis of self-consistency (SC) in long-context scenarios, challenging its presumed universality. Across 651 experiments and eight models, SC consistently \textit{failed to improve performance}---and often degraded it---due to positional biases. Even state-of-the-art models like GPT-4o exhibited these failures, with this degradation persisting across sampling configurations, prompt formats, and evaluation metrics, underscoring SC's fundamental incompatibility with long-context tasks. These findings suggest that inference-time techniques like SC may not generalize to long contexts and that addressing these challenges, particularly the effects of positional bias, will require deeper architectural innovations in attention and aggregation.

For practitioners, our work highlights the risks of porting short-context methods to long-context applications. Future research should investigate hybrid approaches, such as combining SC with positional debiasing, rather than relying on standard SC. By rigorously establishing SC's limitations, this work provides the foundational analysis needed to drive targeted innovations in context-aware inference methods. Rather than attempting to characterize and solve this complex challenge simultaneously, we provide a comprehensive empirical analysis necessary to drive targeted innovations in context-aware inference methods.

\section*{Limitations}
While our study reveals fundamental challenges for SC in long-context settings, several limitations warrant discussion. Our evaluation covers contexts up to 30 documents (5-10K tokens), leaving open questions about extremely long contexts (100K+ tokens) where different dynamics might emerge. However, recent work \citep{modarressi2025nolima} suggests that the effective context window of many models is far shorter than their advertised limits. This finding reinforces our focus as a critical analysis area where models are functional yet exhibit the systematic biases we investigate.

Our focus on textual tasks (summarization, QA, multi-hop reasoning) leaves questions about SC's utility in multi-modal long-context scenarios open. Additionally, while NQ-Open's controlled setup effectively isolates position bias, real-world contexts may exhibit more complex positional interactions than those we captured.

We deliberately focus on problem characterization rather than solution development, which enables more effective future mitigation strategies by establishing a mechanistic understanding of why SC fails. This comprehensive groundwork (651 experiments across multiple models, tasks, and implementations) provides a robust foundation for developing targeted approaches for sampling-based methods for long-context reasoning.

\section*{Acknowledgments}
We would like to thank Asli Celikyilmaz and Minlie Huang, who served as our TACL action editors, and the anonymous reviewers for their comments and feedback, as well as Zhouxiang Fang, Hannah Gonzalez, Emily Guan, Dongwei Jiang, Tianjian Li, Jiefu Ou, Angad Sandhu, Andrew Wang, and Jack Zhang for their constructive discussions which have helped to improve this work. This work was carried out in part at the Advanced Research Computing at Hopkins (ARCH) core facility, which is supported by the National Science Foundation (NSF) grant number OAC1920103.

\bibliography{my, ref}
\bibliographystyle{acl_natbib}

\appendix
\onecolumn
\begin{center}
{\Large \textbf{Supplemental Material}}
\end{center}

\section{Robustness Results}
\label{sec:Results}

\begin{figure*}[ht!]
    \centering
    \resizebox{\width}{0.95\height}{\includegraphics[width=\linewidth]{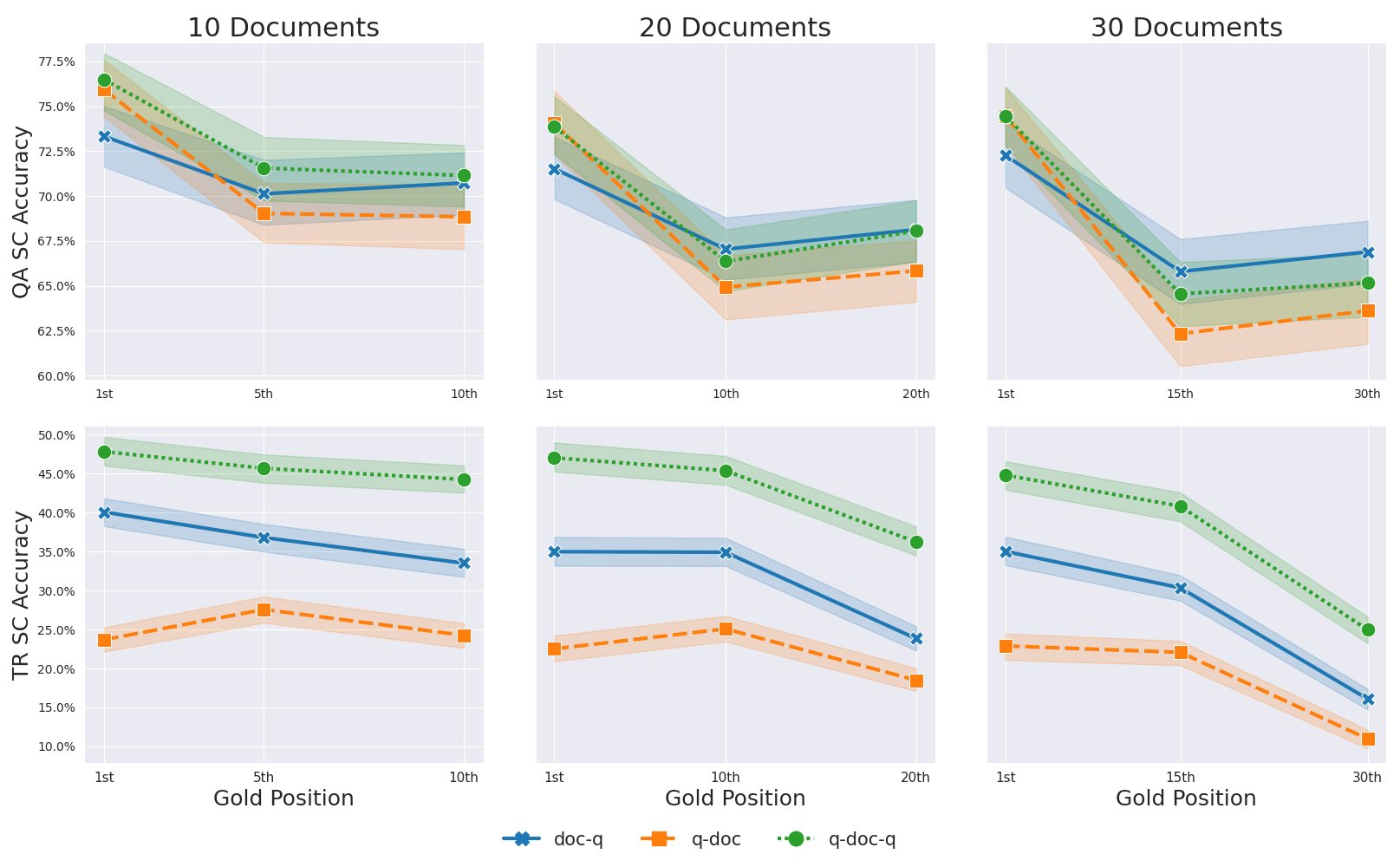}}
    \caption{Effect of prompt format on QA and TR self-consistency accuracy for the Qwen-2.5-7B model. Different prompt formats show minimal impact on mitigating positional bias. While Q-Doc-Q slightly improves overall QA performance (top row), TR performance (bottom row) is more sensitive to format choice, with up to 20\% performance \textit{degradation} between formats. The consistent degradation pattern across gold positions indicates that position bias persists regardless of query-document ordering.}
    \label{fig:qa_prompt}
\end{figure*}

\begin{figure*}[p!]
\centering
    \begin{subfigure}[t]{\linewidth}
        \centering
        \caption{\# of Generations}
        \resizebox{\width}{0.9\height}{\includegraphics[trim={0cm 0.9cm 0cm 0cm},width=\linewidth]{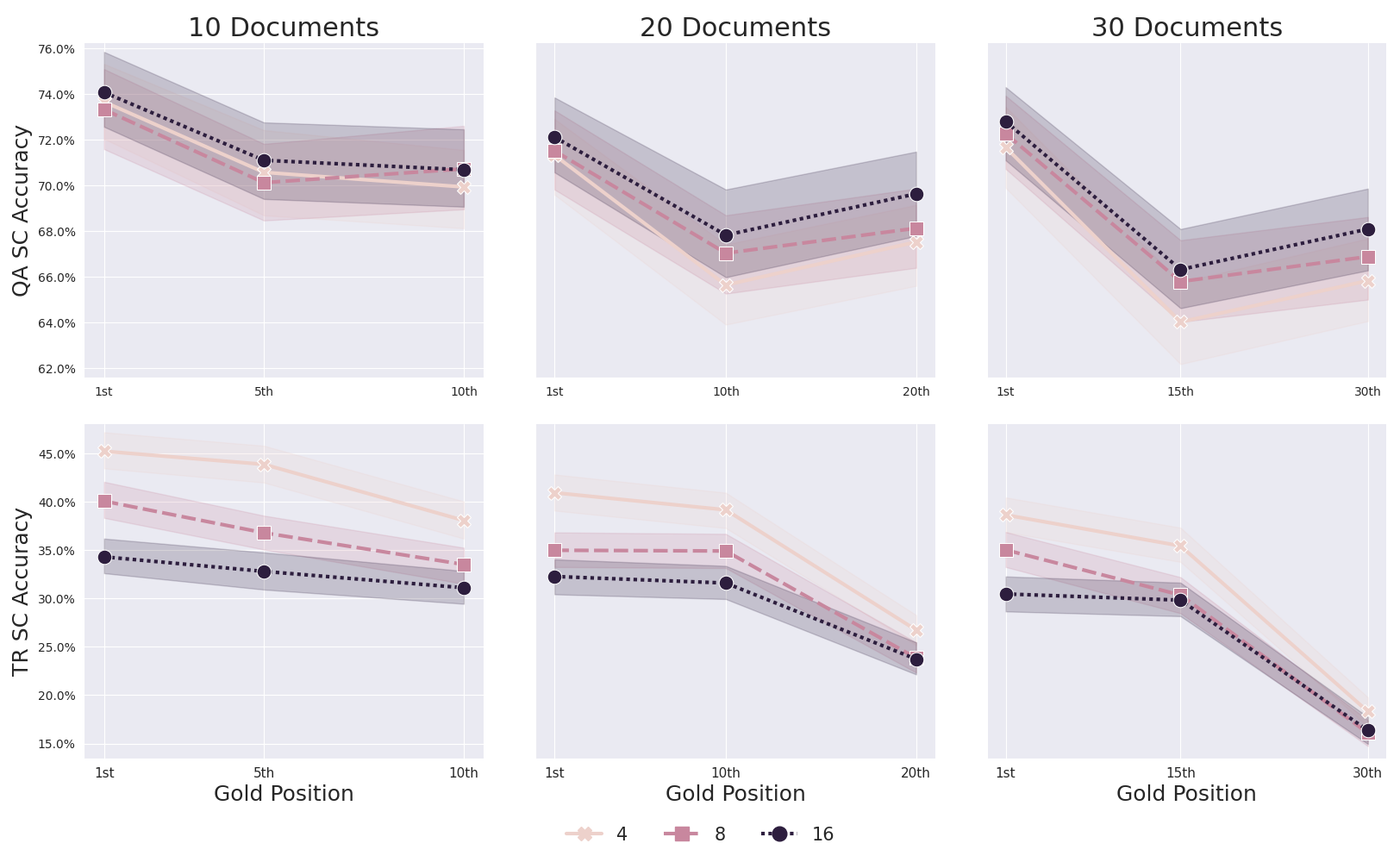}}
        \label{fig:stacked_num_generation}
    \end{subfigure}
    
    \begin{subfigure}[t]{\linewidth}
        \centering
        \caption{Temperature}
        \resizebox{\width}{0.9\height}{\includegraphics[trim={0cm 1.5cm 0cm 0cm},width=\linewidth]{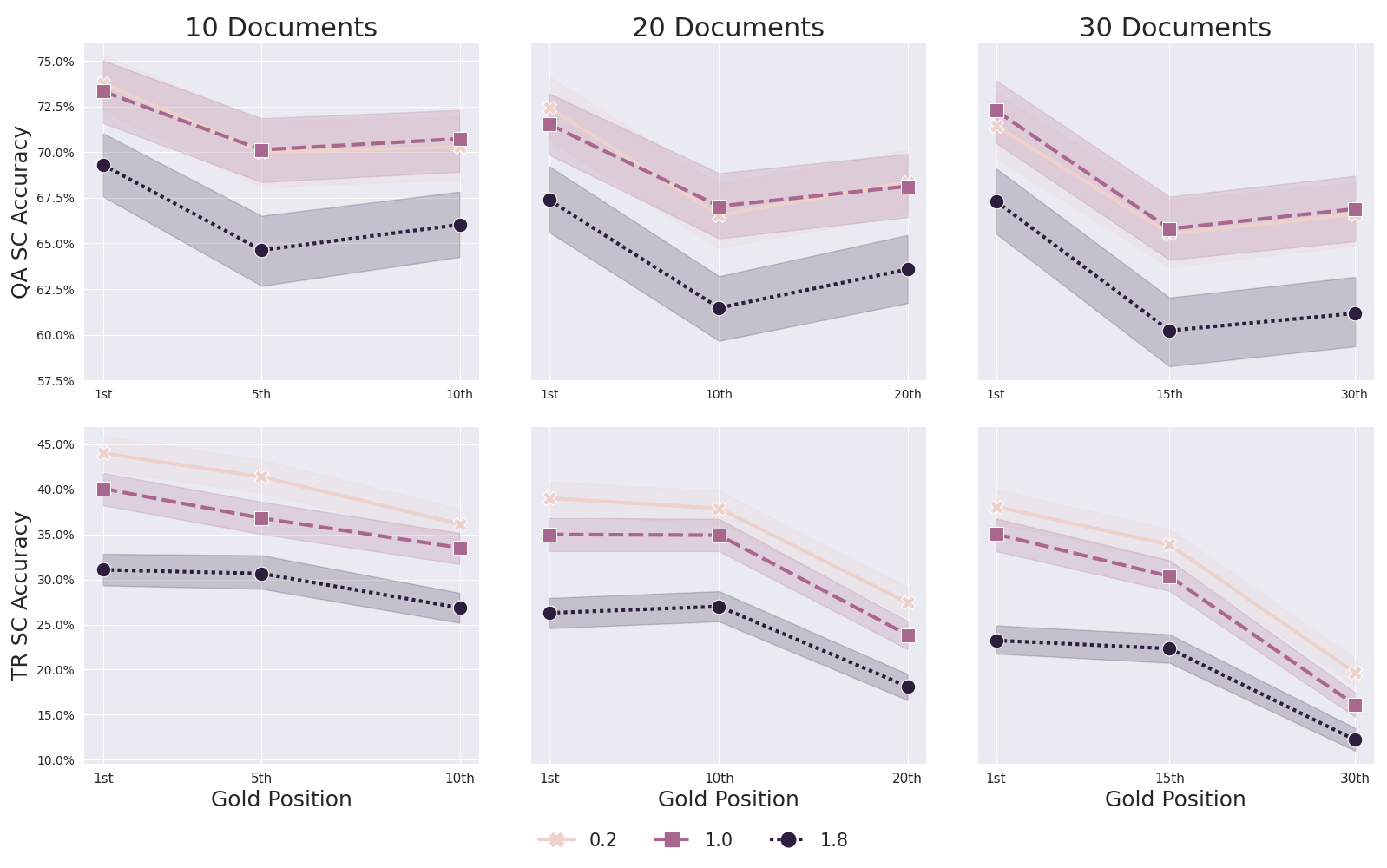}}
        \label{fig:stacked_temperature}
    \end{subfigure}
    \vspace{-0em}
    \caption{Effect of SC parameter variations on QA and TR accuracy for the Qwen-2.5-7B model. Increasing generations (a) slightly boosts QA accuracy but negatively impacts TR. Higher temperatures (b) degrade performance, especially for TR tasks, persisting positional degradation as gold information shifts deeper into the context.}
    \label{fig:stacked_sc_params}
\end{figure*}

\FloatBarrier

\section{Error Analysis of Self-Consistency Failures}
\label{app:ErrorAnalysis}

\begin{figure*}[ht!]
    \centering
    \resizebox{\width}{0.95\height}{\includegraphics[width=\linewidth]{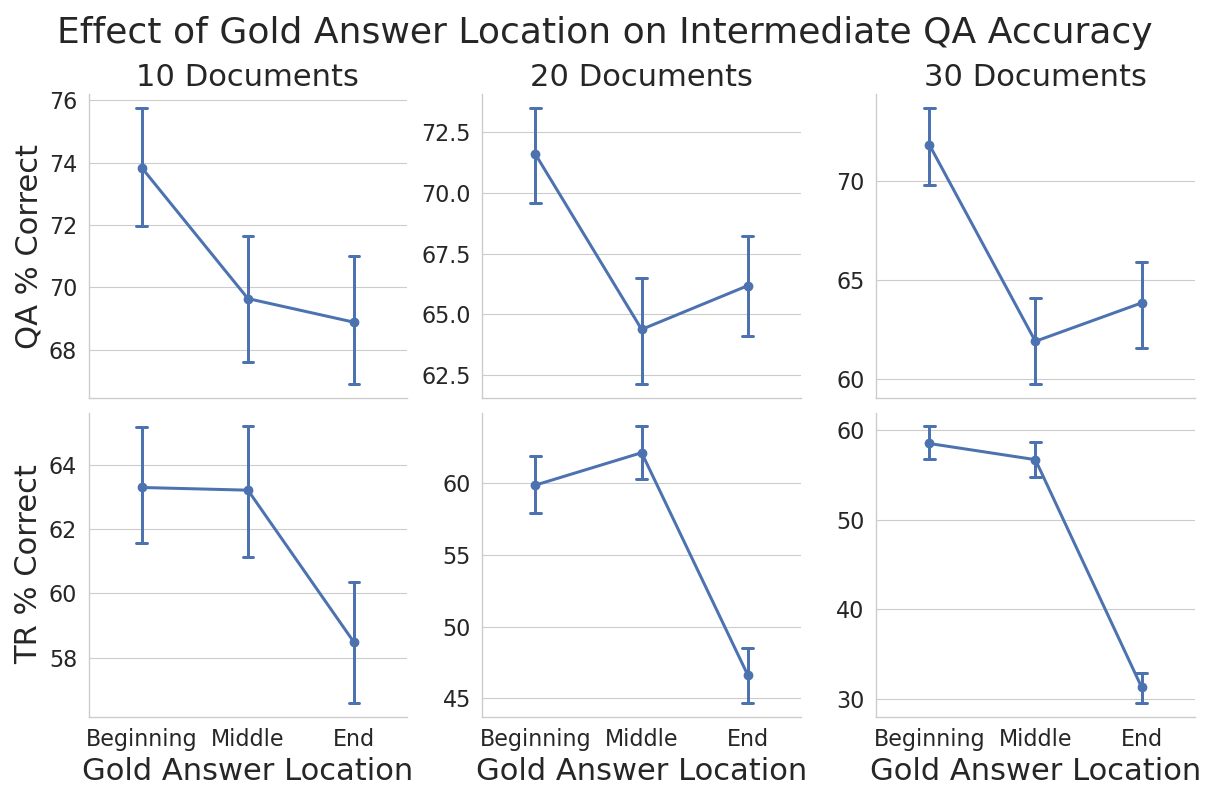}}
    \caption{Error analysis of intermediate generations for the Qwen-2.5-7B model. The accuracy of the eight intermediate samples is plotted before aggregation, revealing the origin of the positional bias observed in Figure~\ref{fig:stacked_qa_tr}. For Question Answering (QA) tasks (top row), accuracy follows a U-shaped curve, degrading in the middle of the context. For Text Retrieval (TR) tasks (bottom row), accuracy shows a monotonic decline as the gold document is placed later in the context.}
    \label{fig:sc_interim_position_bias}
\end{figure*}


\newtcolorbox{examplebox}[1]{
  enhanced,
  breakable, 
  title={#1},
  colback=gray!5,   
  colframe=gray!75, 
  fonttitle=\bfseries,
  coltitle=black,
  colbacktitle=gray!75, 
  attach title to upper, 
  arc=2mm, 
  boxrule=1pt,
}

\begin{examplebox}{Illustrative Example: Correlated Errors in Question Answering}

\vspace{1em}

\textbf{Prompt:} \\
Write a high-quality answer for the given question using only the provided search results (some of which might be irrelevant). \\

Document [1](Title: Drafting of the Universal Declaration of Human Rights) ...\textcolor{red}{John Peters Humphrey} was newly appointed as Director of the Division of Human Rights within the United Nations Secretariat...\\
... \\
\textbf{Document [20](Title: Cyrus Cylinder) The cylinder [of \textcolor{teal}{Cyrus}] has also been referred to as...the first declaration of universal human rights...} \\

Question: who wrote the first declaration of human rights \\
Answer: 

\vspace{1em}

\textbf{Expected Answer:} \\
\texttt{Cyrus}
\end{examplebox}

\begin{examplebox}{Intermediate Generations (8 Samples at Temperature 1.0):}

The model generates eight different reasoning paths. However, due to a systemic position bias, the majority of these paths incorrectly fixate on documents from the beginning of the context, violating SC's core assumption of independent errors.
\begin{enumerate}
    \item \texttt{The first direct quote suggesting the author of the first Declaration of human rights is \textcolor{red}{John Peters Humphrey} can be found in multiple documents.}
    \item \texttt{The first draft of the Universal Declaration of Human Rights, which is often regarded as the first comprehensive declaration of human rights on a global scale, was written by \textcolor{red}{John Peters Humphrey}.}
    \item \texttt{Based on the provided information, specifically from document [8] and document [6], the first draft of what became the Universal Declaration of Human Rights was prepared by \textcolor{red}{John Peters Humphrey}.}
    \item \texttt{The first draft of the Universal Declaration of Human Rights was written by \textcolor{red}{John Peters Humphrey}, a Canadian legal scholar, jurist, and human rights advocate.}
    \item \texttt{Based on the provided documents, \textcolor{red}{John Peters Humphrey} is credited with writing the first draft of the Universal Declaration of Human Rights.}
    \item \texttt{Based on the provided search results, the first draft of the Universal Declaration of Human Rights, often considered a foundational document in human rights, was written by \textcolor{red}{John Peters Humphrey}.}
    \item \texttt{The first explicit declaration of human rights, the Declaration of the Rights of Man and of the Citizen, was set by France's National Constituent Assembly in 1789.}
    \item \texttt{The first declaration of human rights, the Universal Declaration of Human Rights, was primarily drafted by \textcolor{red}{John Peters Humphrey}.}
\end{enumerate}

\vspace{1em}

\textbf{Final Aggregated Answer:} \\
\texttt{\textcolor{red}{John Peters Humphrey}}

\vspace{1em}

\textbf{Analysis:} \\
This example demonstrates the mechanistic failure of self-consistency when faced with strong positional bias. The correct answer, \textbf{\textcolor{teal}{Cyrus}}, is explicitly mentioned in a document near the \textit{end} of the context (Document [20]). However, the model exhibits a powerful primacy bias, causing it to fixate on \textbf{\textcolor{red}{John Peters Humphrey}}, a name found in one of the \textit{first} documents. This bias induces strongly correlated errors across the generated samples, with seven of the eight confidently proposing the same incorrect name. The majority voting process, unable to distinguish this systemic flaw from random noise, latches onto the incorrect consensus and amplifies it.

\end{examplebox}



\section{Prompt Templates}
\label{sec:Prompts}

\newtcolorbox{promptbox}[1]{
  enhanced,
  title={#1},
  colback=gray!5,
  colframe=gray!75,
  fonttitle=\bfseries,
  coltitle=black,
  attach boxed title to top left={yshift=-2mm, xshift=2mm},
  boxed title style={colback=gray!75},
  before skip=1pt,
  after skip=1pt,
  center,
  breakable,
}

\subsection{Summarization Task Prompts}
\label{subsec:SummPrompts}

\begin{promptbox}{GovReport}
You are given a report by a government agency. Write a one-page summary of the report. \\

Report: \\
\{REPORT\} \\

Summary:
\end{promptbox}

\begin{promptbox}{QMSum}
You are given a meeting transcript and a query containing a question or instruction. Answer the query in one or more sentences. \\

Transcript: \\
\{TRANSCRIPT\} \\

Query: \\
\{QUERY\} \\

Answer:
\end{promptbox}

\begin{promptbox}{SQuALITY}
You are given a story and a question. Answer the question in a paragraph. \\

Story: \\
\{STORY\} \\

Question: \\
\{QUESTION\} \\

Answer:
\end{promptbox}

\subsection{Question Answering Task Prompts}
\label{subsec:QAPrompts}

\begin{promptbox}{Qasper}
You are given a scientific article and a question. Answer the question as concisely as you can, using a single phrase or sentence if possible. If the question cannot be answered based on the information in the article, write "unanswerable". If the question is a yes/no question, answer "yes", "no", or "unanswerable". Do not provide any explanation. \\

Article: \\
\{ARTICLE\} \\

Question: \\
\{QUESTION\} \\

Answer:
\end{promptbox}

\begin{promptbox}{NarrativeQA}
You are given a story, which can be either a novel or a movie script, and a question. Answer the question as concisely as you can, using a single phrase if possible. Do not provide any explanation. \\

Story: \\
\{STORY\} \\

Question: \\
\{QUESTION\} \\

Answer:
\end{promptbox}

\begin{promptbox}{QuALITY}
You are provided a story and a multiple-choice question with 4 possible answers (marked by A, B, C, D). Choose the best answer by writing its corresponding letter (either A, B, C, or D). Do not provide any explanation. \\

Story: \\
\{STORY\} \\

Question and Possible Answers: \\
\{QUESTION\_AND\_OPTIONS\} \\

Answer:
\end{promptbox}

\begin{promptbox}{MuSiQue}
You are given several paragraphs from Wikipedia and a question. Answer the question as concisely as you can, using a single phrase if possible. If the question cannot be answered based on the information in the paragraphs, write "unanswerable". Do not provide any explanation. \\

Paragraphs: \\
\{PARAGRAPHS\} \\

Question: \\
\{QUESTION\} \\

Answer:
\end{promptbox}

\subsection{USC Prompt}
\label{subsec:USCPrompts}

\begin{promptbox}{USC}
I have generated the following responses to the question: \{question\} \\

\{RESPONSES\} \\

Evaluate these responses.
Select the most consistent response based on majority consensus.
Start your answer with "The most consistent response is Response X" (without quotes).
\end{promptbox}

\subsection{LLM Judge Prompts}
\label{subsec:JudgePrompts}

\begin{promptbox}{LLM Judge (Summarization)}
You are an expert evaluator assessing the quality of a generated summary compared to a reference summary. Consider the following factors in your evaluation: \\
- Consistency: Does the generated summary convey the same key information as the reference summary without contradictions? \\
- Relevance: Is the generated summary focused on the main points of the reference summary? \\
- Fluency: Is the generated summary grammatically correct and free of repetition or incoherence? \\
- Informativeness: Does the generated summary adequately cover the most important details in the reference? \\

Provide your evaluation as a single score on a scale from 0 to 100: \\
- 0 means the generated summary is completely unacceptable (e.g., incoherent, irrelevant, or contradictory). \\
- 100 means the generated summary is perfect (e.g., consistent, relevant, fluent, and informative). \\

Reference Summary: \\
\{REFERENCE\} \\

Generated Summary: \\
\{PREDICTION\} \\

Score (0-100):
\end{promptbox}

\vspace{1em}

\begin{promptbox}{LLM Judge (QA)}
You are an expert evaluator assessing the quality of an answer generated in response to a question. Consider the following factors in your evaluation: \\
- Correctness: Does the generated answer accurately address the question based on the ground truth? \\
- Relevance: Is the generated answer focused and does it address the question without unnecessary information? \\
- Fluency: Is the generated answer grammatically correct and free of repetition or incoherence? \\

Provide your evaluation as a single score on a scale from 0 to 100: \\
- 0 means the generated answer is completely unacceptable (e.g., incorrect, irrelevant, or incoherent). \\
- 100 means the generated answer is perfect (e.g., correct, relevant, and fluent). \\

Question: \\
\{QUESTION\} \\

Ground Truth Answer: \\
\{REFERENCE\} \\

Generated Answer: \\
\{PREDICTION\} \\

Score (0-100):
\end{promptbox}

\subsection{Position Bias Analysis Prompts}
\label{subsec:PositionBiasPrompts}

\subsubsection{Task Prompts}
\label{app:TaskTypes}

\begin{promptbox}{QA Task:}
\{DOCUMENTS\} \\

\textbf{Question:} \{QUESTION\} \\
Answer: 
\end{promptbox}

\vspace{1em}

\begin{promptbox}{TR Task:}
\{DOCUMENTS\} \\

\textbf{Which documents are needed to answer the following query:} \{QUESTION\} \\
Answer: 
\end{promptbox}

\vspace{1em}

\subsubsection{Prompt Formats}
\label{app:PromptFormats}

\begin{promptbox}{DOC-Q:}
\{DOCUMENTS\} \\

\textbf{Question:} \{QUESTION\} \\
Answer: 
\end{promptbox}

\vspace{1em}

\begin{promptbox}{Q-DOC:}
\textbf{Question:} \{QUESTION\} \\

\{DOCUMENTS\} \\

Answer: 
\end{promptbox}

\vspace{1em}

\begin{promptbox}{Q-DOC-Q:}
\textbf{Question:} \{QUESTION\} \\

\{DOCUMENTS\} \\

\textbf{Question:} \{QUESTION\} \\
Answer: 
\end{promptbox}

\end{document}